\documentclass[journal,letterpaper]{IEEEtran}
\usepackage[numbers,sort]{natbib}

\usepackage{amsmath,amsfonts}
\usepackage{array}
\usepackage{textcomp}
\usepackage{stfloats}
\usepackage{url}
\usepackage{verbatim}
\usepackage{graphicx}
\usepackage{xspace}
\newcommand{\astar}{A*\xspace}  % just in case we want to change how it's formatted e.g. A\textsuperscript{$\ast$}

\usepackage{float}                  % Allows 'Here and Only Here' [H] for Floats
\usepackage{url}                    % \url{} command
\usepackage{graphicx}               % \includegraphics
\usepackage{amsmath}
\usepackage{amsthm}
\usepackage{booktabs}
\usepackage{multirow}
\usepackage{tikz}
\usetikzlibrary{positioning}
\usetikzlibrary{fit,shapes}
\usetikzlibrary{calc}
\usetikzlibrary{matrix, positioning,fit,shapes,calc}
\graphicspath{{../../../src/},{../../../src/},{../../../img/},{../../img/},{../../../../img/},{../img/},{img/},,{../../../}}
\usepackage{algorithm}
\usepackage{algpseudocode}

\makeatletter
\newcounter{phase}[algorithm]
\newlength{\phaserulewidth}
\newcommand{\setphaserulewidth}{\setlength{\phaserulewidth}}

\makeatother

\setphaserulewidth{.7pt}

\usepackage{subcaption}
\usepackage{adjustbox}
\usepackage[english]{babel} 
\usepackage{varioref} 
\usepackage[bookmarks=true,bookmarksopen=true,breaklinks]{hyperref} 
\usepackage[nameinlink]{cleveref}
\hypersetup{
  colorlinks   = true,              %Colours links instead of ugly boxes
  urlcolor     = black,              %Colour for external hyperlinks
  linkcolor    = black,              %Colour of internal links
  citecolor    = black,                %Colour of citations
  breaklinks=true   % splits links across lines
}
\usepackage[hyphenbreaks]{breakurl}

\crefname{table}{table}{tables}
\Crefname{table}{Table}{Tables}
\crefname{figure}{figure}{figures}
\Crefname{figure}{Figure}{Figures}
\crefname{equation}{equation}{equations}
\Crefname{equation}{Equation}{Equations}
\addto\extrasenglish{  
    
}
\addto\extrasenglish{  
    
}
\addto\extrasenglish{  
    
}
\addto\extrasenglish{  
    
}

\addto\extrasenglish{  
    
}

\addto\extrasenglish{  
    
}
\addto\extrasenglish{  
    
}
\addto\extrasenglish{  
    
}

\addto\extrasenglish{  
    
}
\algnewcommand{\LineComment}[1]{\State \(\triangleright\) #1}

\algnewcommand{\LeftComment}[1]{\(\triangleright\) #1}
\algnewcommand{\TestLineComment}[1]{// #1}

\newcommand{\refappendix}[1]{\hyperref[#1]{Appendix}}

\makeatletter
\newcommand*{\centerfloat}{%
  \parindent \z@
  \leftskip \z@ \@plus 1fil \@minus \textwidth
  \rightskip\leftskip
  \parfillskip \z@skip}
\makeatother

\makeatletter
\newcommand\Autoref[1]{\@first@ref#1,@}
\def\@throw@dot#1.#2@{#1}% discard everything after the dot
\def\@set@refname#1{%    % set \@refname to autoefname+s using \getrefbykeydefault
    \edef\@tmp{\getrefbykeydefault{#1}{anchor}{}}%
    \xdef\@tmp{\expandafter\@throw@dot\@tmp.@}%
    \ltx@IfUndefined{\@tmp autorefnameplural}%
         {\def\@refname{\@nameuse{\@tmp autorefname}s}}%
         {\def\@refname{\@nameuse{\@tmp autorefnameplural}}}%
}
\def\@first@ref#1,#2{%
  \ifx#2@\autoref{#1}\let\@nextref\@gobble% only one ref, revert to normal \autoref
  \else%
    \@set@refname{#1}%  set \@refname to autoref name
    \@refname~\ref{#1}% add autoefname and first reference
    \let\@nextref\@next@ref% push processing to \@next@ref
  \fi%
  \@nextref#2%
}
\def\@next@ref#1,#2{%
   \ifx#2@ and~\ref{#1}\let\@nextref\@gobble% at end: print and+\ref and stop
   \else, \ref{#1}% print  ,+\ref and continue
   \fi%
   \@nextref#2%
}

\makeatother

\makeatletter
\renewcommand*{\NAT@spacechar}{~}
\makeatother
\usepackage{microtype}
\makeatletter
\newenvironment{lflalign*}{%                                                         
  \def\align@preamble{%                                                             
    &\strut@                                                                        
    \setboxz@h{\@lign$\m@th\displaystyle{####}$}%                                   
    \ifmeasuring@\savefieldlength@\fi                                               
    \set@field                                                                      
    \hfil                                                                           
    \tabskip\z@skip                                                                 
    &\setboxz@h{\@lign$\m@th\displaystyle{{}####}$}%                                
    \ifmeasuring@\savefieldlength@\fi                                               
    \set@field                                                                      
    \hfil                                                                           
    \tabskip\alignsep@                                      
  }                                                                                 
\start@align\tw@\st@rredtrue\m@ne
  }
{\endalign}
\makeatother
\title{Towards Objective Metrics for Procedurally Generated Video Game Levels}

\author{Michael Beukman,
Steven James and Christopher Cleghorn \thanks{M. Beukman, S. James and C. Cleghorn are with the School of Computer Science and Applied Mathematics, University of the Witwatersrand. This work has been submitted to the IEEE for possible publication. Copyright may be transferred without notice, after which this version may no longer be accessible.}
}

\begin{document}

\maketitle
\begin{abstract}
    With increasing interest in procedural content generation by academia and game developers alike, it is vital that different approaches can be compared fairly. However, evaluating procedurally generated video game levels is often difficult, due to the lack of standardised, game-independent metrics. In this paper, we introduce two simulation-based evaluation metrics that involve analysing the behaviour of a planning agent to measure the diversity and difficulty of generated levels in a general, game-independent manner. Diversity is calculated by comparing action trajectories from different levels using the edit distance, and difficulty is measured as how much exploration and expansion of the A* search tree is necessary before the agent can solve the level.
    We demonstrate that our diversity metric is more robust to changes in level size and representation than current methods and additionally measures factors that directly affect playability, instead of focusing on visual information. 
    The difficulty metric shows promise, as it correlates with existing estimates of difficulty in one of the tested domains, but it does face some challenges in the other domain.
    Finally, to promote reproducibility, we publicly release our evaluation framework.\footnote{\url{https://github.com/Michael-Beukman/PCGNN}}
\end{abstract}

\section{Introduction} % Main chapter title
\label{chap:intro} % For referencing this chapter elsewhere, use \ref
\IEEEPARstart{P}{rocedural} Content Generation (PCG) is a large field~\citep{summerville2018procedural_pcgml,togelius2011searchbased} with many different approaches, ranging from genetic algorithms~\citep{ferreira_mario} to reinforcement learning~\citep{pcgrl}.
A major problem, however, is the lack of comparable metrics between different games and works of literature. 
For example, \textit{leniency}~\citep{smith2010launchpad_leniency} measures the difficulty of levels based on atomic challenges (e.g., jumping over a gap in a platformer), which must be defined separately for each new game~\citep{shaker2012evolving} and might not apply to games without a concept of danger to the player, such as puzzle games. 

We address this problem by introducing two simulation-based metrics that only use the behaviour of a planning agent on a level to judge its difficulty and diversity. In particular, we use \astar~\citep{hart1968formal}, but any other planning algorithm would suffice. Difficulty is measured based on how much the agent needs to explore before solving the level, and the diversity between two levels is calculated by comparing the action trajectories obtained from each level. These metrics require no game-specific feature engineering or domain knowledge---simply a game engine and an agent.

We test our methods on a \textit{Maze} game as well as \textit{Super Mario Bros}. 
Our results indicate that our diversity metric displays desirable characteristics, such as invariance to visual changes that do not affect playability, as well as being less sensitive to the size or representation of the tested levels than the metrics we compare against. The difficulty metric correlates with existing measures in one of the tested domains, but fails to do so in the other.
\section{Related Work}
\label{chapter:background}
How one chooses to evaluate generated levels for comparison between different methods is important in PCG. Many evaluation schemes, however, are mostly ad hoc and not easily transferable to other scenarios. 
For example, \citet{pcgrl} use preset ``goal'' criteria to evaluate the level. These criteria stipulate that there should be a lower bound on the number of steps needed to solve a level, and that specific objects must occur a certain number of times (e.g. there must be 1 player, 1 door, etc.).
\citet{constrained_novelty}, who use an evolutionary approach, mainly report population-based metrics, such as how many individuals in the final population were feasible (i.e. playable and valid according to the game engine), and how diverse the feasible population is. Diversity between two tilemaps here is measured as the average number of non-matching tiles.
\citet{ferreira_mario} simply state that their method generates levels that are similar to the original \textit{Super Mario Bros.} levels, without defining a similarity metric.
It is therefore difficult to compare these results, as they usually measure different aspects, and often do not use the same units or methods of measuring.

It would be useful to have a consistent method for evaluating a level that potentially includes metrics such as diversity, difficulty, and quality. \citet{horn2014comparative} echo this sentiment and attempt to create a valid range of metrics that can be used to evaluate levels, but these have not been widely adopted. Most of the metrics are defined for a single level, measuring aspects like \textit{linearity} (how well a level can be described by a straight line), \textit{density} (how many platforms are stacked on top of each other in platform games), and \textit{leniency} (how easy a level is). 
One metric that compares the diversity of levels is the compression distance, which measures the relative similarity between two levels from the same generator. The authors add that promising future work could be to use a simulation-based evaluation score, where an agent plays the generated level, and its behaviour is used to score the level.

Evaluating procedurally generated content is thus a major challenge, since different works adopt different game-specific approaches, preventing fair comparisons and hindering progress in the field as a whole.
\section{Existing and Proposed Metrics}
\label{chap:metrics}
We now describe existing metrics and present our new simulation-based metrics.
Our proposed metrics are general and evaluate the \textit{diversity} and \textit{difficulty} of generated levels without requiring game-specific configuration or human interaction. We compare them to the compression distance and leniency metrics respectively.

\subsection{Existing Metrics}
\subsubsection{Compression Distance}
Compression distance (CD) is a metric that measures the similarity between two strings by determining how much space is saved by compressing them as one concatenated string, compared to compressing them separately~\citep{li2004similarity}. In the case of PCG, we simply use a string representation of the levels under consideration.

The \textit{normalised compression distance} (NCD) is defined as
$$\text{NCD}(x, y) = \frac{C(xy) - min\{ C(x), C(y)  \}}{max\{ C(x), C(y) \}}$$
where $C(x)$ is the length of the string $x$ after compressing it, and $xy$ is simply the concatenation of $x$ and $y$. The intuition is that strings that are very similar will have a low NCD value as compressing them together can take advantage of their similarities. On the other hand, strings that are very different will have a higher distance, as compressing them as one string does not offer much benefit over compressing them separately. For clarity, we refer to this metric (using gzip as the compression algorithm) simply as CD.

\citet{shaker2012evolving} apply this to platformer levels by generating a feature for every column in the map, which is determined by various aspects, like an increase or decrease in platform height, the start or end of a gap, the existence of enemies and other blocks, as well as their combinations. This string of features is then used as the string representation of a level.
For the \textit{Maze} Game, we simply flatten the map, and thereby obtain a binary string that represents the level, where zeros and ones represent free space and walls respectively.

\subsubsection{Leniency}
Leniency is a metric that measures how lenient a level is to a player's mistakes. Originally defined for platformers by \citet{smith2010launchpad_leniency}, it was determined by averaging the \textit{leniency score} for each of the potential challenges in a level, and normalising that to a value between 0 and 1. These challenges could be a gap or an enemy (with a leniency value of $l=-1$), a jump with no associated gap ($l=1$), and various other elements. We follow the approach of \citet{shaker2012evolving}, who use leniency in the context of \textit{Super Mario Bros.} by slightly adapting the metric described above.

For the \textit{Maze} game, it is not immediately obvious how to apply this metric, since there are no inherent challenges. We therefore elect to calculate leniency as the fraction of paths that result in dead ends. For each unfilled tile $t$, we find the shortest path from the start $s$ to $t$. We then fill in this path and determine if there is still a path from $t$ to the goal $g$. Tile $t$ is labelled as a dead end only if no such path exists. 

\subsection{\astar Diversity}
We assume that diverse levels require diverse solutions~\citep{diversity_is_all_you_need}, and we thus look at the differences in solution characteristics between two levels to calculate their diversity.

To measure the diversity between two levels, we simply run our \astar agent separately on each level, and consider the actions that the agent performed as integers specifying a single discrete action (like moving to the right in the \textit{Maze} game). We use these action strings to evaluate how different the levels are, using the Levenshtein (or edit) distance~\citep{levenshtein1966binary}, so that trajectories of different lengths can be compared.
We normalise this distance by dividing it by the length of the longer string.

This diversity metric measures levels that can be solved using closely-related approaches as \textit{similar}, and other levels that have drastically different paths as \textit{diverse}. This method thus focuses on the characteristics of a level that actually affect the playability, instead of being distracted by different (but potentially irrelevant) visual information.

There might be some potential problems with this metric, however. Specifically, we use actions to distinguish levels, and some actions might not affect the environment, such as walking into a wall~\citep{diversity_is_all_you_need}. This is mitigated by using an agent that rarely performs unnecessary actions. A second possible shortcoming is that multiple actions might result in roughly the same path (e.g., moving to the right in \textit{Super Mario Bros.} is often equivalent to jumping right), leading to noise in the metric. Again though, using the same agent in all cases leads to consistent and comparable actions.

We further compute the correlation between this edit distance-based metric and a similar one that uses the average Manhattan distance between corresponding positions, which should be less affected by small differences in actions. We find a high correlation (Pearson's $r > 0.7, p < 0.05$), indicating that this phenomenon does not have a large effect.

Finally, we re-emphasise that this method is not constrained to only use \astar, and it can be used with any game-playing agent, such as an evolutionary controller or reinforcement learning agent.

\subsection{\astar Difficulty}
We measure the difficulty of a level as the number of nodes that the \astar agent expands that are not along the optimal path, divided by the total number of reachable states.
The justification for this is that levels that are more difficult and contain more dead ends will require more exploration, and thus more expansion of the search tree. 
On the other hand, easier levels, where there is a simple unobstructed path between the start and the goal, will require much less searching, as the heuristic will lead the agent directly to the goal.

\section{Results}
\begin{figure*}
    \centerfloat
    \resizebox{1\textwidth}{!}{%
    \includegraphics[height=10cm]{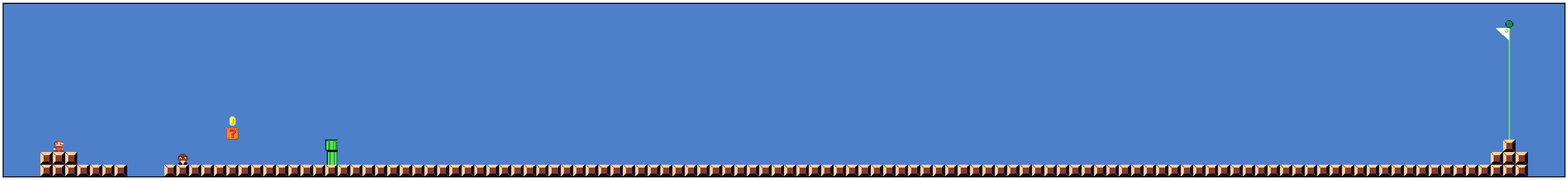}%
    \quad
    \fbox{\includegraphics[height=10cm]{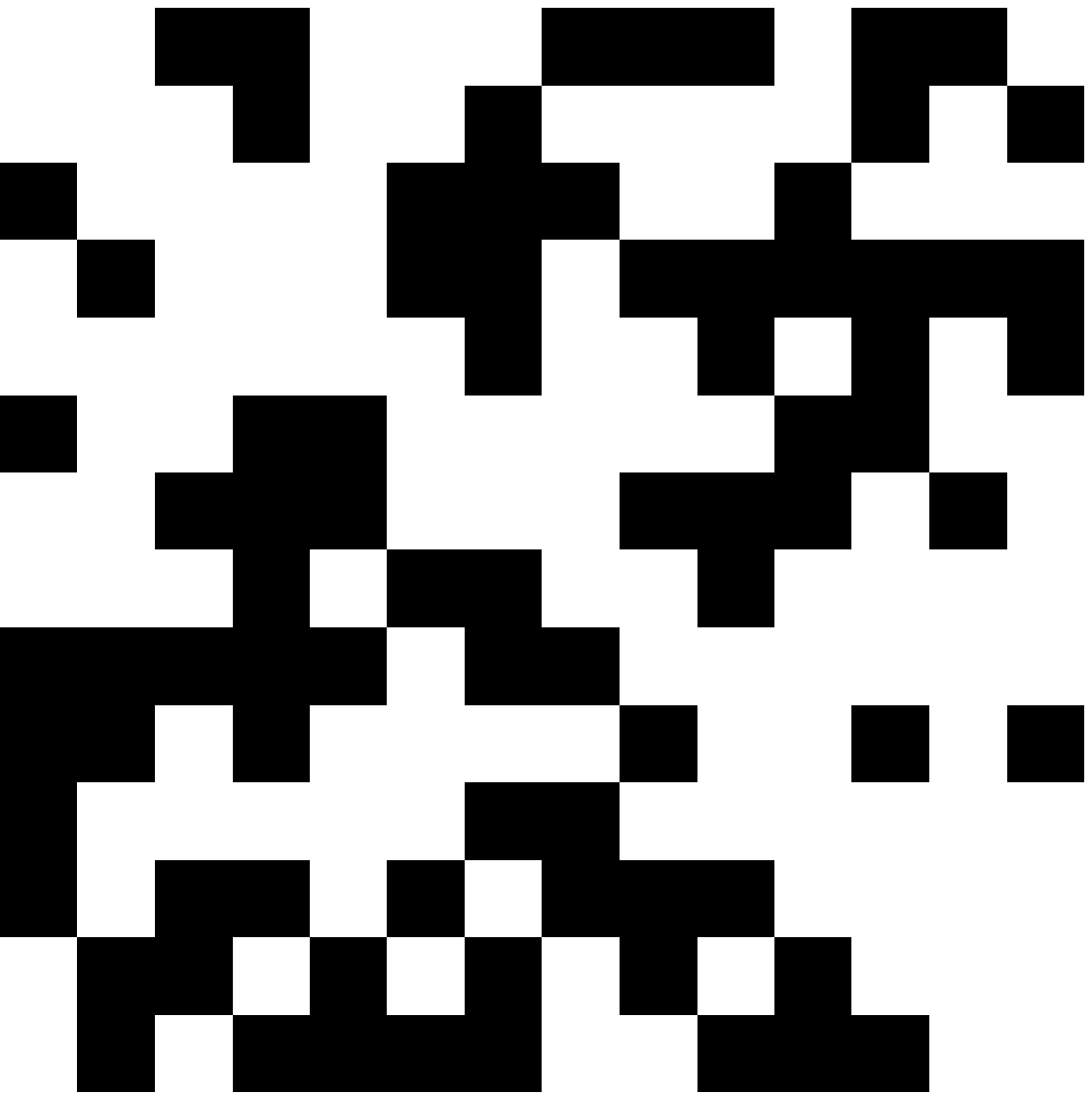}}%
    }

    \caption{Example levels for the \textit{Super Mario Bros.} (left) and \textit{Maze} (right) domains. In \textit{Super Mario Bros.}, the agent must traverse the screen to reach the flag at the rightmost edge, while in the \textit{Maze} domain, the agent must find a path from the top left to the bottom right.}
    \label{fig:example_levels_method}
\end{figure*}

We investigate two tilemap-based games. The first is a simple \textit{Maze} game where each tile can be a wall or empty space, and the goal is to find a path between the top left corner and the bottom right corner. We use the Manhattan distance to the goal as the \astar heuristic here. The second game we consider is a simplified version of \textit{Super Mario Bros.}, with only Goombas as enemies, and no powerups. We use the \astar agent by Robin Baumgarten\footnote{\url{https://github.com/amidos2006/Mario-AI-Framework/tree/master/src/agents/robinBaumgarten}}~\citep{togelius20102009_with_baumgarten}, where the heuristic is based on the estimated time to reach the rightmost point of the screen. Example levels can be seen in \autoref{fig:example_levels_method}.

We evaluate levels generated using a variety of methods: PCGRL~\citep{pcgrl}, a genetic algorithm approach~\citep{ferreira_mario}, and a method that uses NeuroEvolution of Augmenting Topologies~\citep{neat_without_doi} combined with novelty search~\citep{novelty_search} to evolve level generators. This latter method was also used to generate levels of different sizes, as it proved the fastest and generated mostly solvable levels.

\subsection{Diversity}
\label{sec:metrics_diversity_sec}
When comparing these two diversity metrics, we find a few specific characteristics and major flaws of compression distance:
\begin{enumerate}
    \item Compression distance is sensitive to the exact string representation used, and there is a marked difference between simply flattening the \textit{Super Mario Bros.} levels and using a more game-specific feature representation.
    \item The compression distance increases with the size of the level, while the variance decreases. Larger levels are thus all marked as having approximately the same diversity score, regardless of their actual content.
\end{enumerate}

To illustrate these shortcomings, we consider the distribution of diversity scores, i.e. if we have 100 levels, there are $\frac{(100)(99)}{2} = 4950$ pairwise comparisons, and we consider all of these values for 5 different random seeds.

\begin{figure}[H]
    \centerfloat
    \includegraphics[width=1\linewidth]{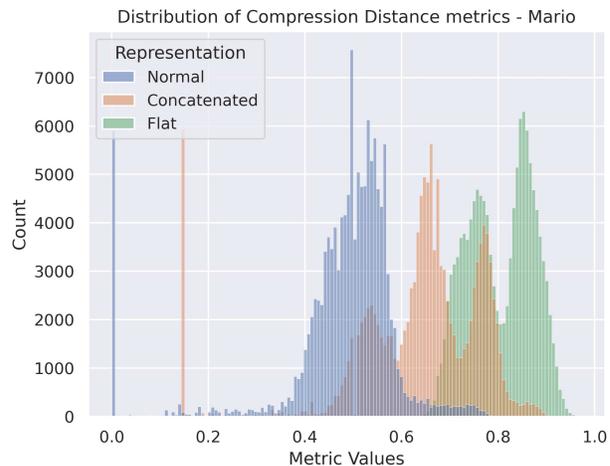}
    \caption{Distributions of compression distance when using 3 different string representations of the \textit{Super Mario Bros.} domain. This showcases a high sensitivity to an irrelevant factor---the specific representation used.}
    \label{fig:cd_string_repr}
\end{figure}

\autoref{fig:cd_string_repr} shows that the compression distance is quite sensitive to the exact representation used. We consider three different representations: firstly ``Normal'', as described above and detailed by~\citep{shaker2011feature,shaker2012evolving}. Secondly, we use ``Concatenated'', which is similar to Normal, but we instead concatenate the strings representing the platform height and the enemy placement respectively~\citep{shaker2011feature}. Finally, we consider ``Flat'', where we simply flatten the 2D array of tiles.
We find that the correlation between these methods, especially for the Flat representation, is not very large. Normal and Concatenated are better correlated, however, as they are related. Our method, however, does not suffer from this problem, as it is completely independent of the representation used.

\begin{figure}
    \centerfloat
    \begin{subfigure}[t]{1\linewidth}
        \includegraphics[width=1\linewidth]{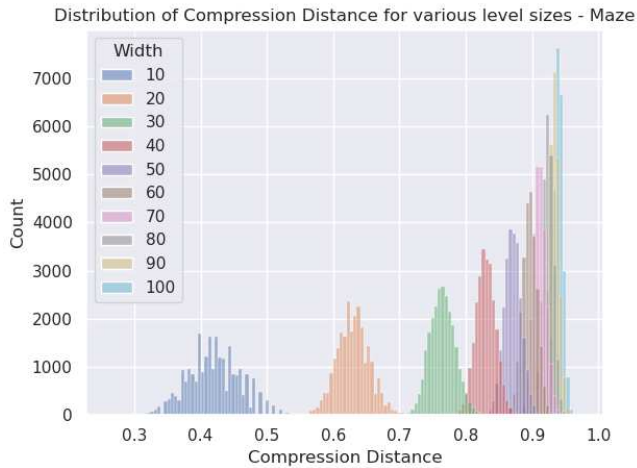}
        \subcaption{The compression distance values become increasingly meaningless as the level size increases.}
        \label{fig:cd_distro_maze}
    \end{subfigure}
    \\
    \centerfloat
    \begin{subfigure}[t]{1\linewidth}
        \includegraphics[width=1\linewidth]{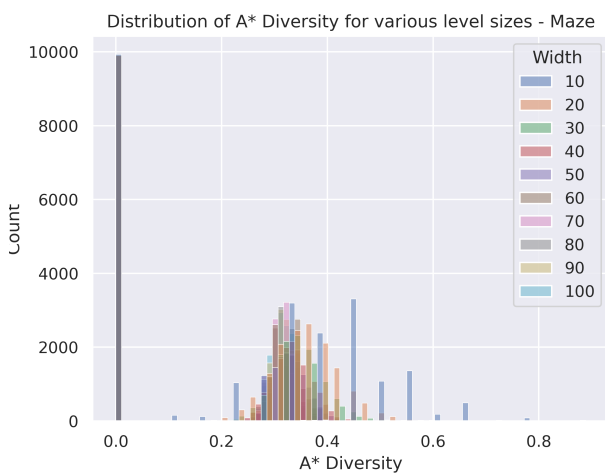}
        \subcaption{By contrast, the \astar Diversity is not as affected by the level size.}
        \label{fig:astar_div_distro_maze}
    \end{subfigure}
    \caption{Distributions of diversity metrics for the \textit{Maze} domain. Each colour represents the metric values from pairwise comparisons over 5 seeds and 100 levels per seed for different level sizes. Only solvable levels that were generated using one specific method were considered (to fairly compare against the \astar metric, which requires solvability), but the compression distance trend also holds for random levels.}
    \label{fig:diversity_distro_maze}
\end{figure}

\begin{figure*}
    \centerfloat
    \begin{subfigure}[t]{0.45\linewidth}
        \includegraphics[width=1\linewidth]{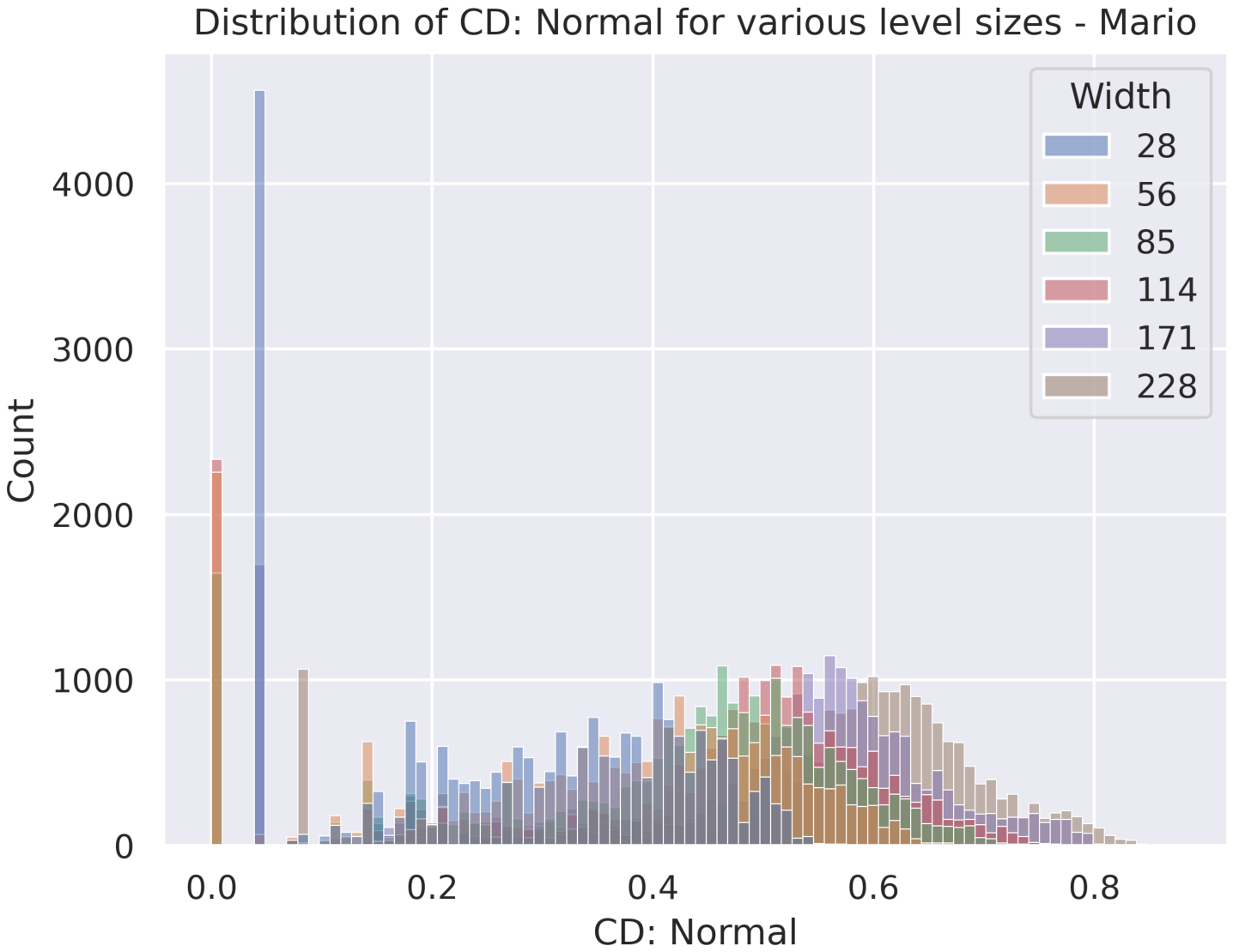}
        \caption{}
    \end{subfigure}
    \begin{subfigure}[t]{0.45\linewidth}
        \includegraphics[width=1\linewidth]{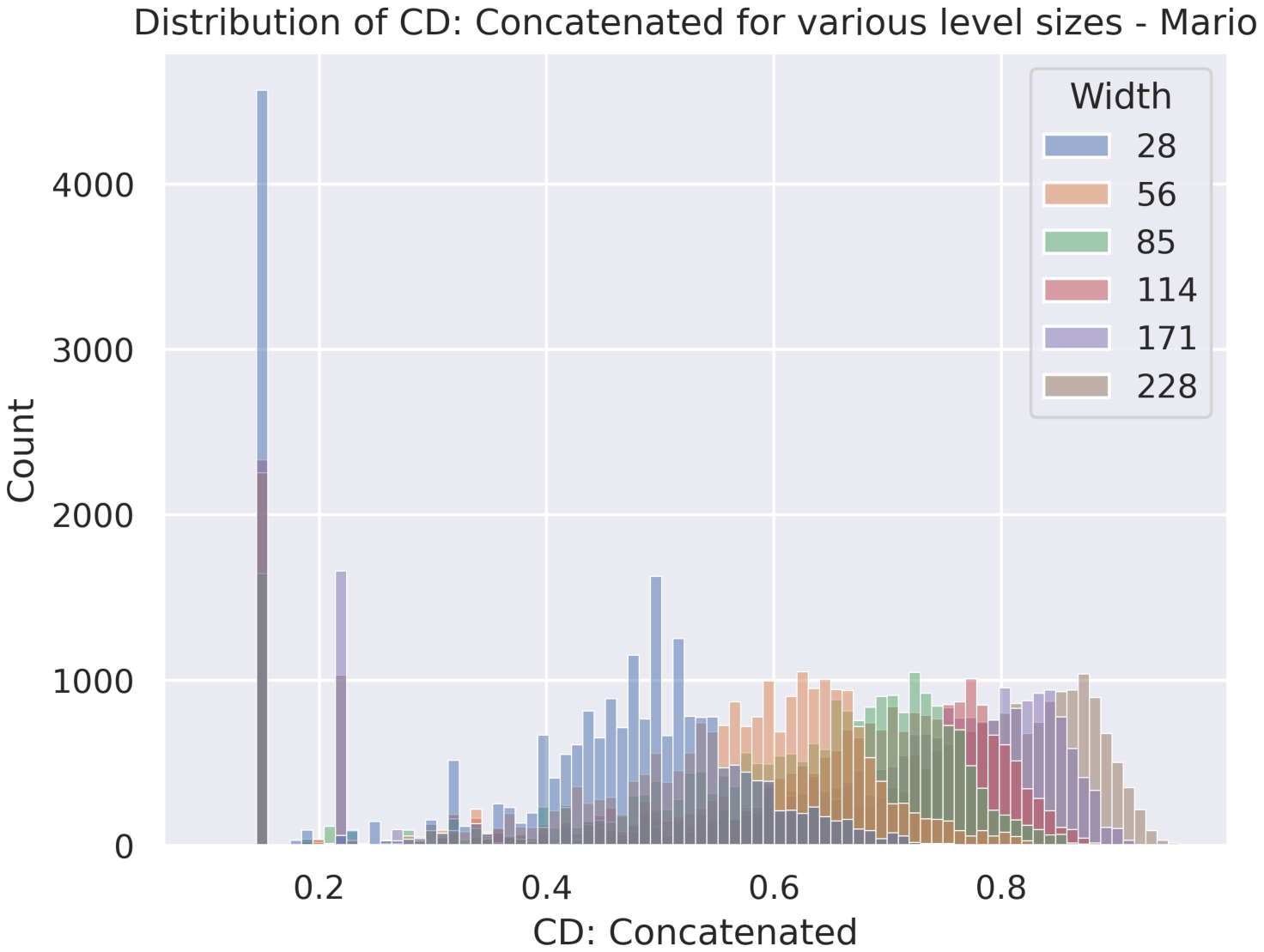}
        \caption{}
    \end{subfigure}
    \\
    \begin{subfigure}[t]{0.45\linewidth}
        \includegraphics[width=1\linewidth]{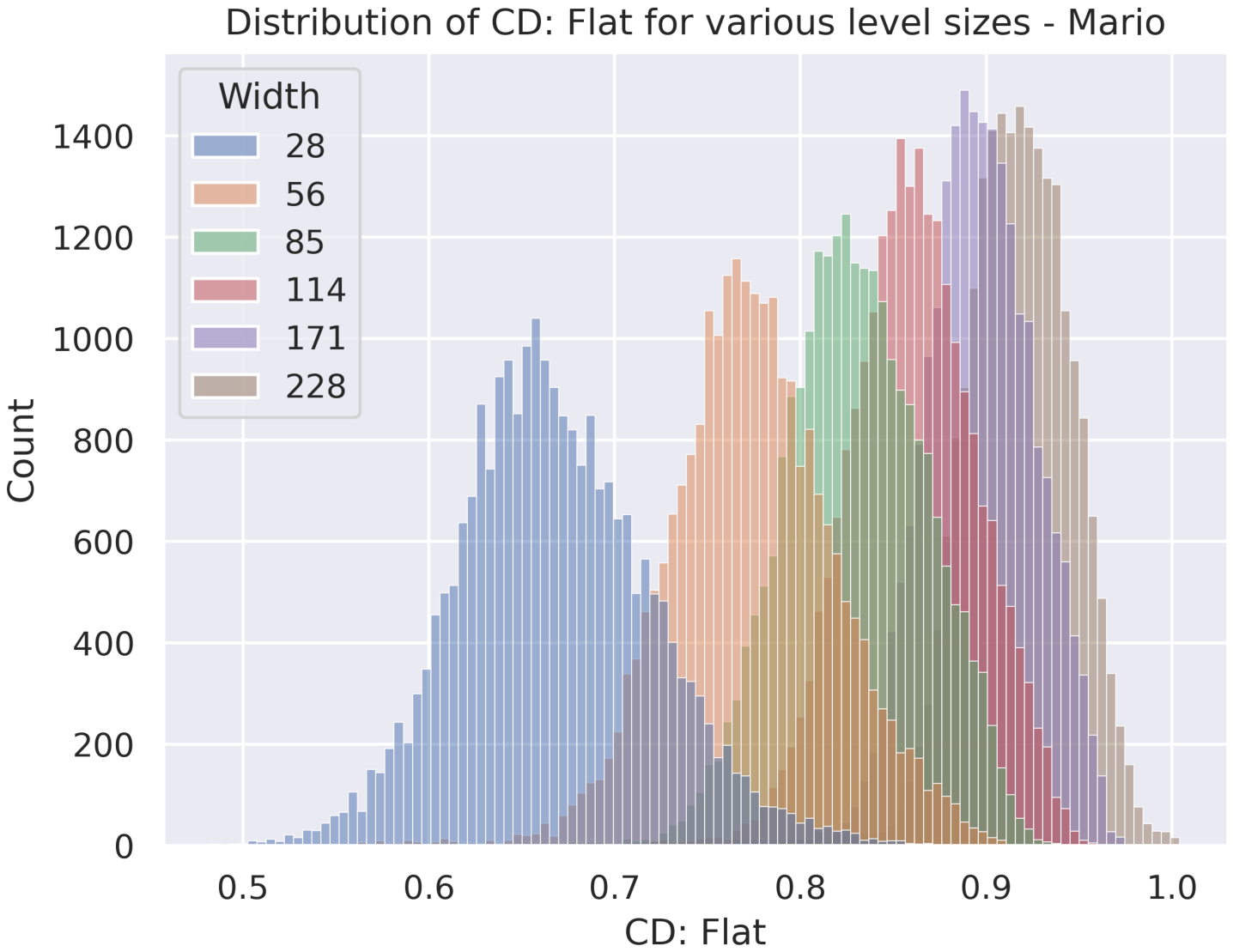}
        \caption{}
    \end{subfigure}
    \begin{subfigure}[t]{0.45\linewidth}
        \includegraphics[width=1\linewidth]{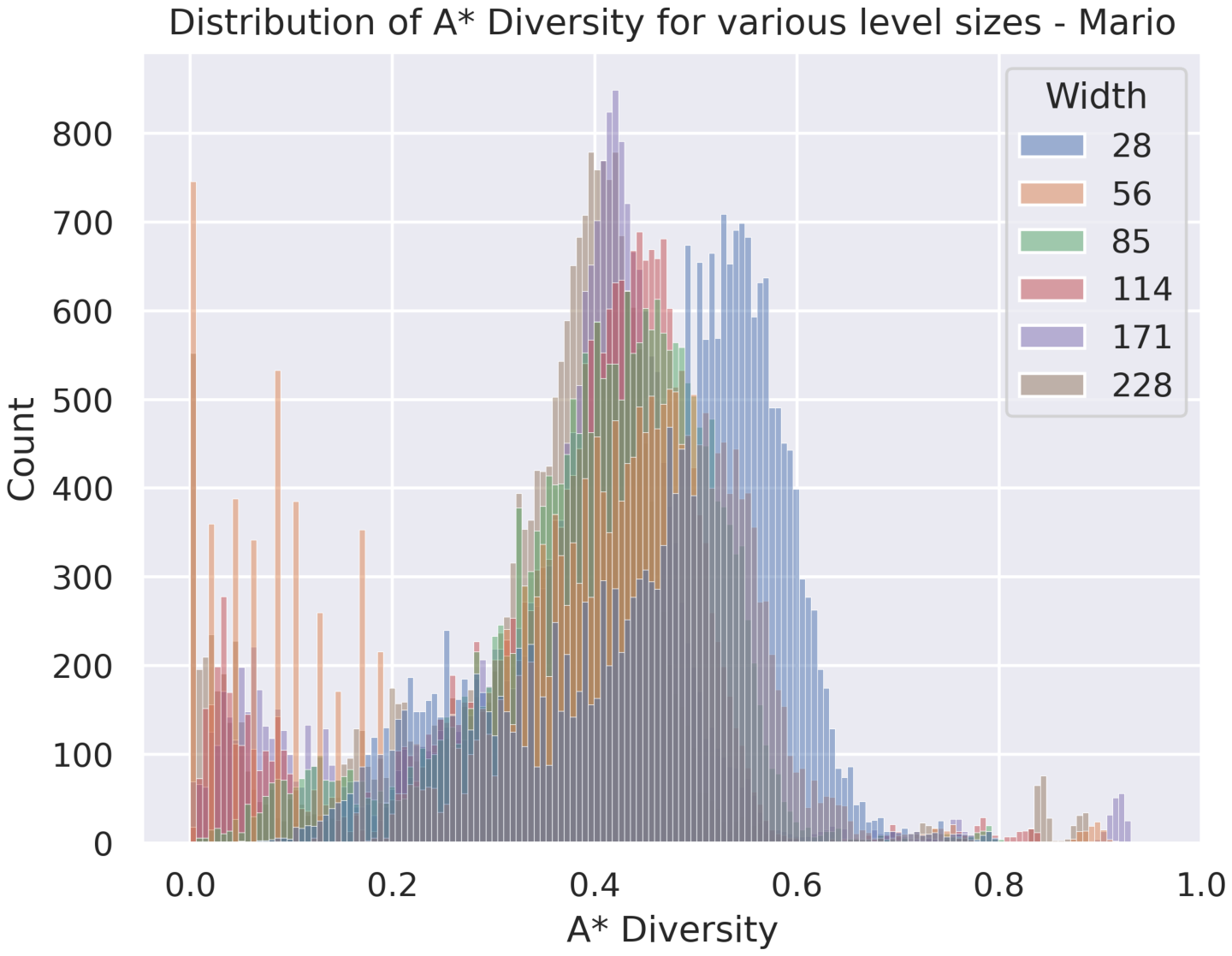}
        \caption{}
    \end{subfigure}
    \caption{Illustrating the diversity metrics as level size increases in the \textit{Super Mario Bros.} domain. (a--c) For all level representations, the compression distance displays sensitivity to the level size. Further, for a fixed size of levels, the different representations display drastically different distributions, mirroring \autoref{fig:cd_string_repr}'s results. (d) Our diversity metric is robust to increases in level size.}
    \label{fig:diversity_distro_mario}
\end{figure*}

\autoref{fig:diversity_distro_maze} shows that as the level size increases, the mean of the compression distance becomes larger, while its variance reduces. By contrast, the \astar diversity metric shows less dramatic changes when the level size increases. We also note that our metric marks many pairs of levels as being identical (i.e. \astar $\text{diversity}=0$), as the same set of actions solved both levels, which is something that the compression distance does not do. We see a similar effect for \textit{Super Mario Bros.} in \autoref{fig:diversity_distro_mario}, but it is much more pronounced when using the flat representation.

\begin{figure}
    \centering
    \begin{subfigure}[t]{0.32\linewidth}
        \fbox{\includegraphics[width=1\linewidth]{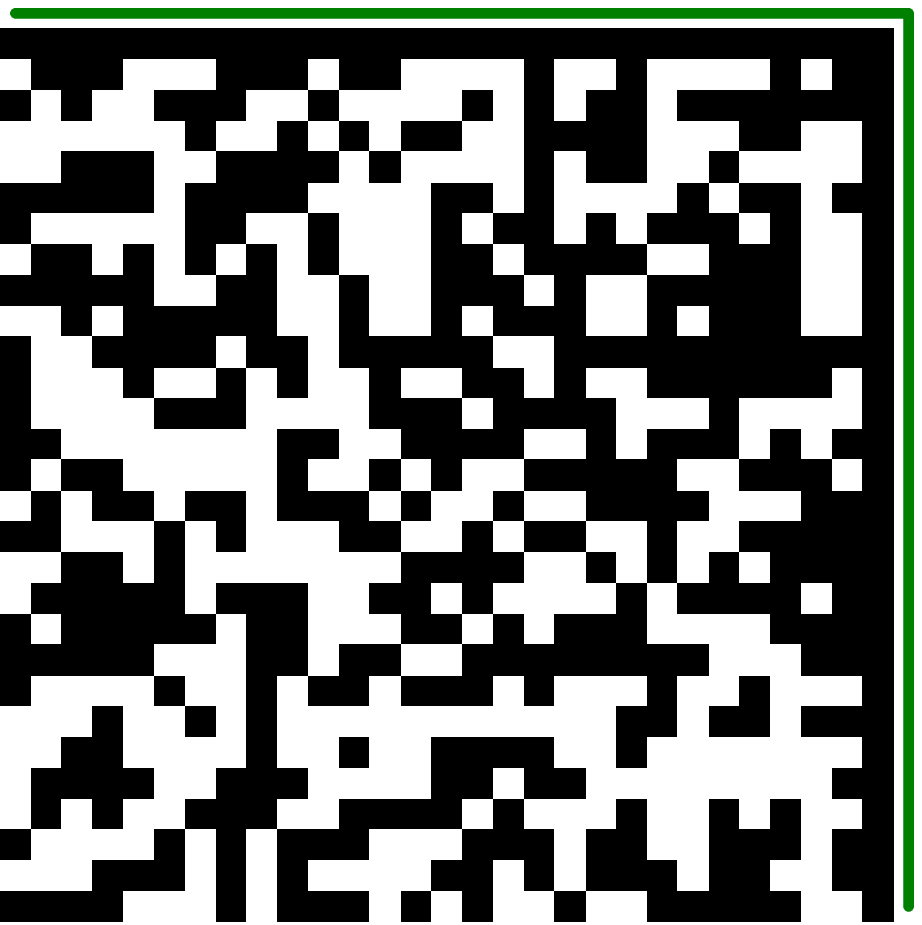}}
    \end{subfigure}
    \begin{subfigure}[t]{0.32\linewidth}
        \fbox{\includegraphics[width=1\linewidth]{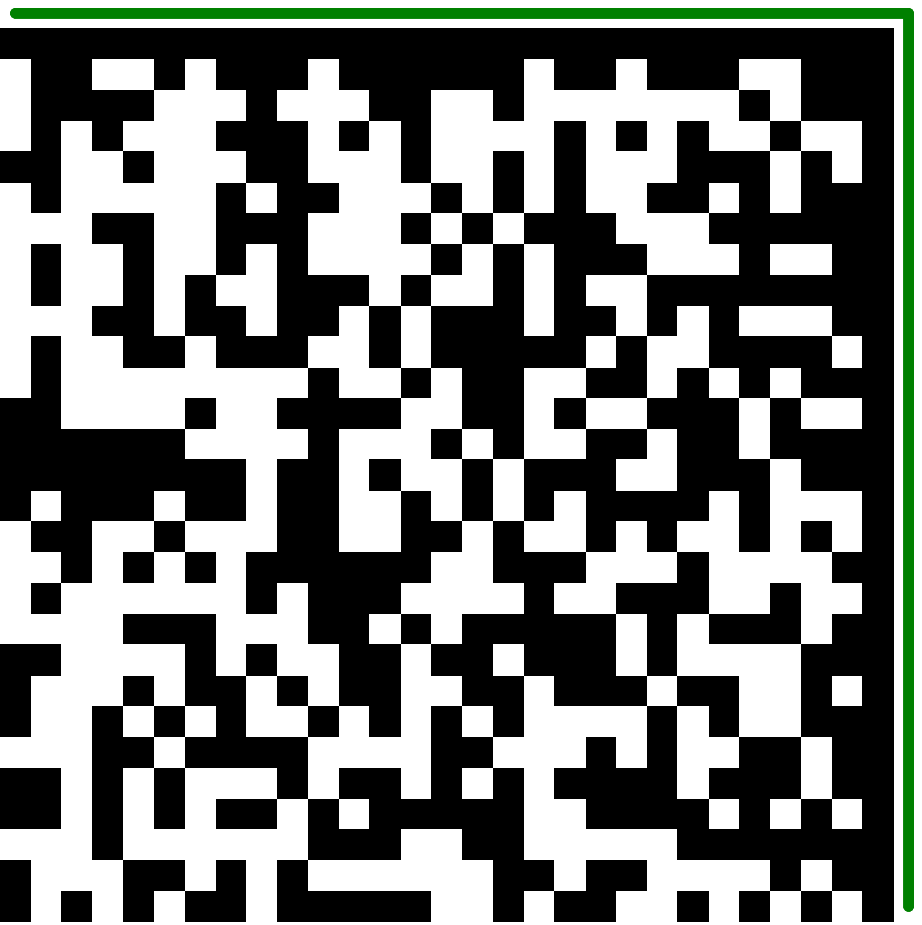}}
    \end{subfigure}
    \begin{subfigure}[t]{0.32\linewidth}
        \fbox{\includegraphics[width=1\linewidth]{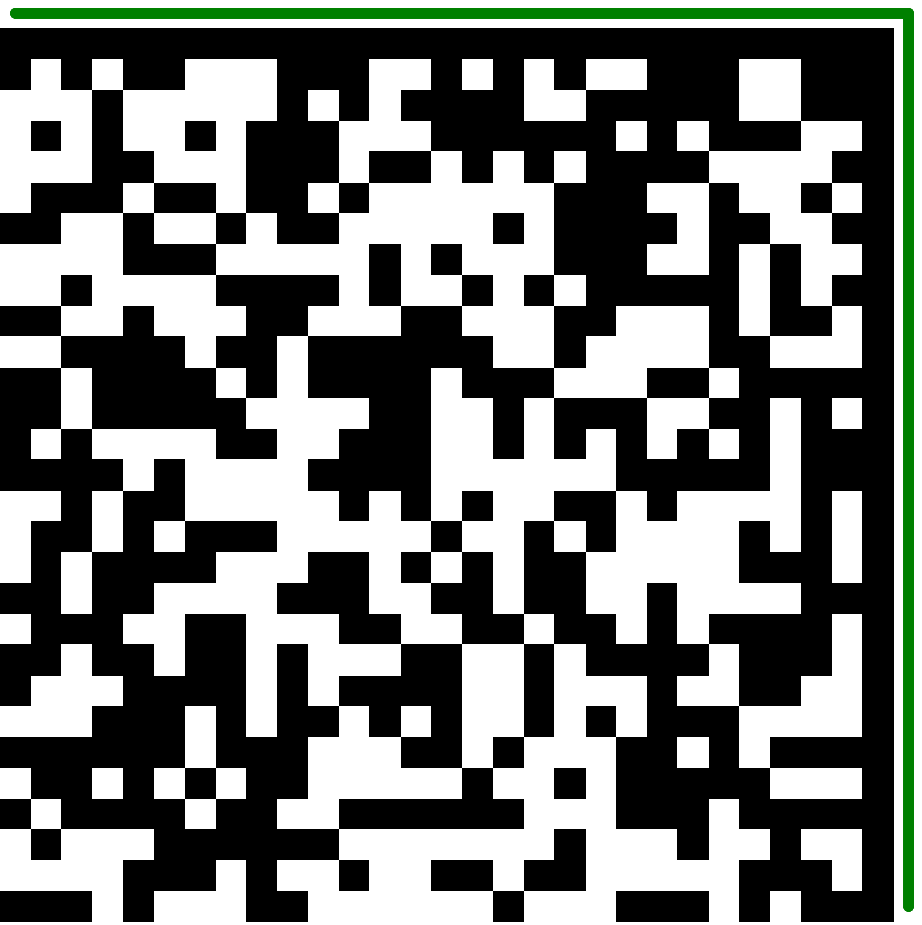}}
    \end{subfigure}
    \caption{Example levels that are visually similar while playing identically. The solution  path (shown in green) is identical for all levels.}
    \label{fig:cd_visual_changes_examples}
\end{figure}

\begin{figure}
    \centering
    \includegraphics[width=0.8\linewidth]{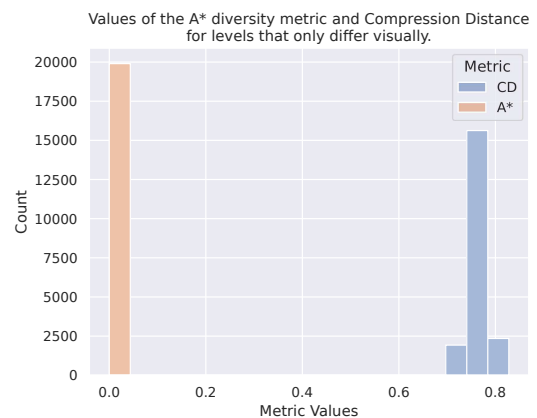}
    \caption{Showcasing the diversity metric values for levels that only differ visually in the \textit{Maze} domain. Compression distance rates all levels as diverse, despite the fact that they are functionally identical.}
    \label{fig:cd_visual_changes}
\end{figure}

Finally, to illustrate that compression distance also considers visual information that does not affect the playing experience of the level, we generate a set of $30\times 30$ levels with one unchanging path from the start to the end. The rest of the tiles, which are unreachable, are randomly set. A few of these levels can be seen in \autoref{fig:cd_visual_changes_examples}, with results shown in \autoref{fig:cd_visual_changes}. Compression distance does not make a distinction between arbitrary visual differences and differences in playable area, and thus assigns most pairs of levels high diversity, whereas the \astar diversity metric marks all levels as identical.

Thus, compression distance has some undesirable characteristics that make it less suitable as a comparable, objective metric. Concretely, it is very dependent on the specific string representation used, as well as the size of the levels under consideration. By contrast, the \astar diversity metric is independent of the level representation used, and is less affected by changes such as level size.

\subsection{Difficulty}

We next analyse the \astar difficulty metric, comparing it to the preexisting leniency metric. 
When comparing the leniency of a collection of levels with their \astar difficulty score, we find a Pearson's correlation coefficient of $-0.17$ ($p = 3.4 \times 10^{-30}$) over $\pm 4000$ levels for the \textit{Maze} , and we find no statistically significant correlation over $\pm 2500$ levels for \textit{Super Mario Bros}. 
We next investigate whether there exists a correlation between the original \textit{Super Mario Bros.} levels and the metrics above; in particular, whether later levels imply a higher difficulty (lower leniency) than earlier levels. We use the implementation\footnote{\url{http://sokath.com/fdg2014_pcg_evaluation/}} from \citet{horn2014comparative} to measure the leniency of a subset of 10 of the original levels, but we find no statistically significant correlation between the level index and difficulty for either leniency or the \astar metric. This might indicate that later levels are not necessarily more difficult than previous ones, but rather just different due to, for example, new enemies or features. During this experiment, we also found that some levels were not solvable by the \astar agent, since it usually took a greedy path to get to the rightmost point of the screen, and some levels had barriers that needed to be surpassed with backtracking.

We next perform a similar experiment on the \textit{Maze} domain using mazes generated according to a desired difficulty,\footnote{\url{http://www.glassgiant.com/maze/}} which we treat as a ground-truth label. We seek to determine whether our metric is able to distinguish easy mazes from harder ones by generating 20 levels (of size $40 \times 40$) for each of the 5 difficulty categories ranging from ``very easy'' to ``very difficult''.

\begin{figure}
    \centering
    \begin{subfigure}{0.49\linewidth}
        \includegraphics[width=1\linewidth]{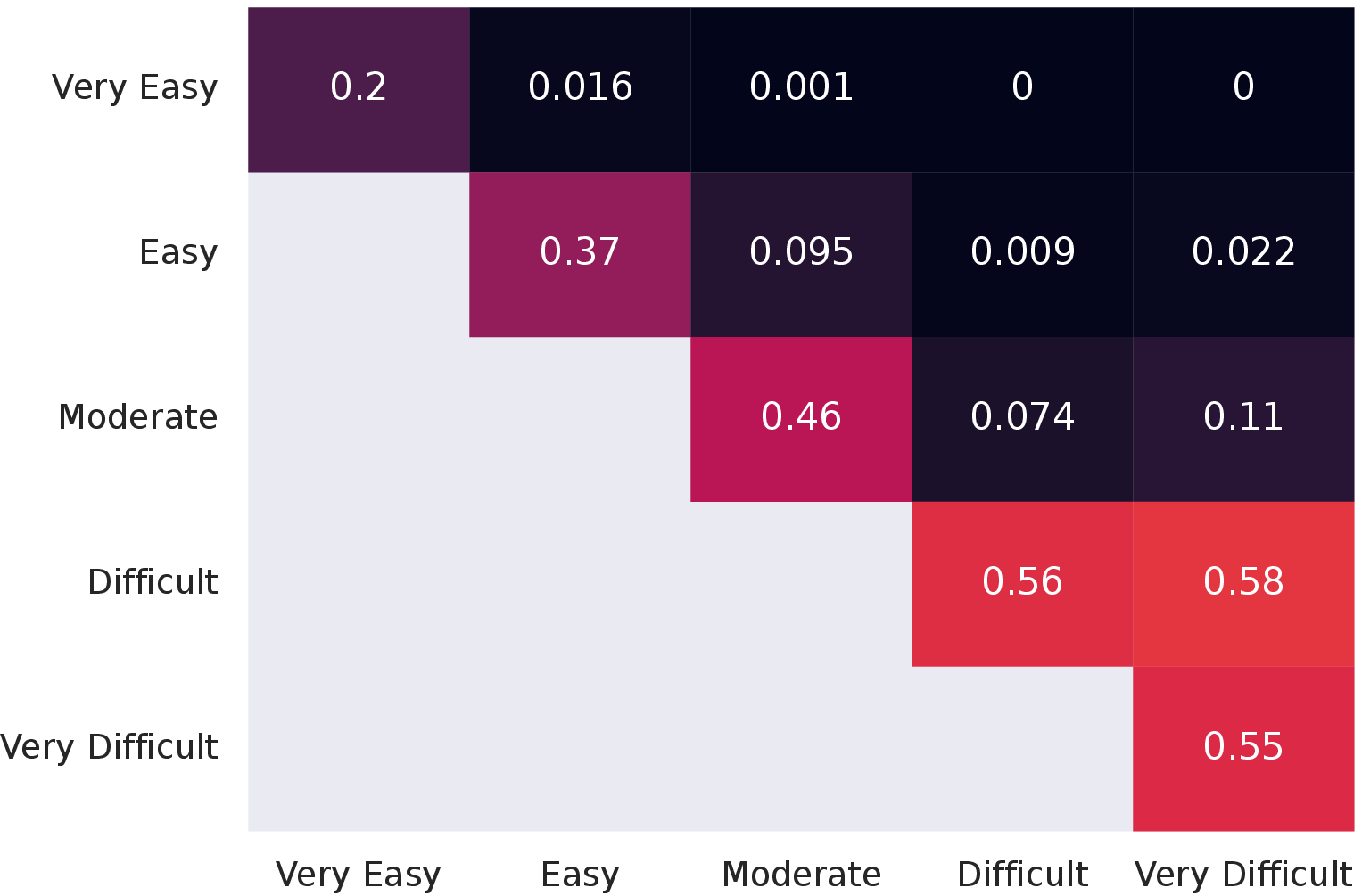}
        \subcaption{\astar Difficulty}
    \label{fig:difficulty_glassgiant}
    \end{subfigure}
    \begin{subfigure}{0.49\linewidth}
        \includegraphics[width=1\linewidth]{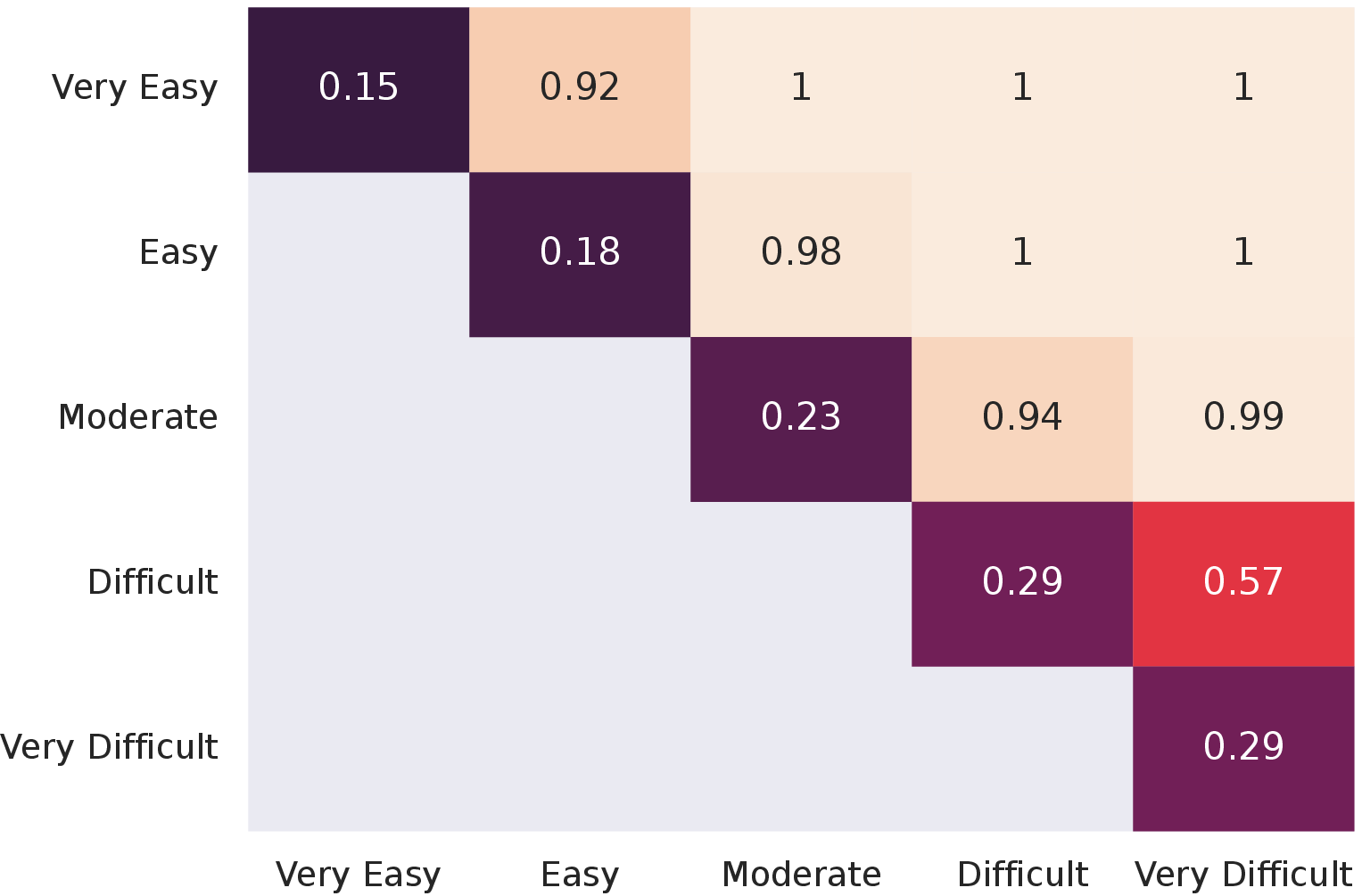}
    \subcaption{Leniency}
    \label{fig:difficulty_glassgiant_leniency}
    \end{subfigure}
    \caption{Comparing difficulty from different classes. The diagonal elements are the average metric values for that category. The off-diagonal elements $(i, j)$ are the $p$-values when obtained from a one-sided Mann-Whitney U test~\citep{mann_whitney_test} that compares the scores from class $i$ and $j$. The alternative hypothesis is that \astar($i$) $<$ \astar($j$) in (a) and  $\text{Leniency}(i) > \text{Leniency}(j)$ in (b) (as being less lenient indicates higher difficulty). The results indicate that the \astar difficulty finds roughly the correct ordering of difficulty, whereas leniency does not.}
    \label{fig:all_glassgiant_diff}
\end{figure}

The results are shown in \autoref{fig:all_glassgiant_diff}, where we observe a general increase in \astar difficulty as  ground-truth difficulty increases, although ``moderate'', ``difficult'' and ``very difficult'' are marked as similar by our metric. The $p$-values indicate that ``very easy'' and ``easy'' levels indeed have lower \astar difficulty scores than ``moderate'', ``difficult'' and ``very difficult''.
By contrast, the leniency score increases with the ground-truth difficulty. This is unexpected, as a higher leniency actually implies a \textit{lower} difficulty, and suggests that the leniency measure does not always correspond to human notions of difficulty.

To determine whether these results generalise, we perform the same experiment using a different maze generator.\footnote{\url{https://github.com/mwmancuso/Personal-Maze}}
While noisier, the results still indicate a similar trend in that the easiest difficulty is rated by the \astar difficulty metric to be easier than the hardest one. The results for leniency are similar to what is shown in \autoref{fig:difficulty_glassgiant_leniency}, although there is much less variance in the leniency scores, with all difficulties, except for the most difficult one, being marked as similarly lenient.

\section{Discussion, Future Work and Conclusion}

\label{chap:conclusion}

\label{sec:disussion}

We introduce two general agent-based metrics that measure the diversity and difficulty of levels. These metrics do not require any game-specific knowledge or intricate feature representations, but simply a game engine and an agent. Our diversity metric is more expressive than compression distance for the \textit{Maze} game, and is not as dependent on the size of the levels, or the string representation used.

Although preliminary results demonstrate that our difficulty metric based on the the state tree of an \astar agent correlates with existing difficulty estimates for the \textit{Maze} domain, more work is needed to verify whether this does indeed capture the general notion of ``difficulty''.
This could be achieved by correlating its scores with the time taken for humans to complete a given set of levels.

Promising avenues for future work also include comparing the \astar diversity metric to the KL-Divergence based score introduced by \citet{edrl} or against human judgements of diversity.
A learning-based approach could also be used to measure difficulty, where the difficulty is proportional to how long an agent (e.g. a reinforcement learning agent) needs to learn before being able to solve a level~\citep{evolution_assesment_sokoban}. 
Different types of agents can even be used to approximate different levels of player skill~\citep{gonzalez2020finding}.

Overall, we believe that the metrics proposed here are a step towards standardising the evaluation of procedural content generation---an important step in accelerating research in the field.

\section*{Acknowledgements}
This work is based on the research supported wholly by the National Research Foundation of South Africa (Grant UID 133358). Computations were performed using High Performance Computing infrastructure provided by the Mathematical Sciences Support unit at the University of the Witwatersrand.

\bibliography{bib}

\end{document}